\ifdefined\pdfminorversion
\fi

\PassOptionsToPackage{dvipsnames}{xcolor}
\documentclass[sigconf,screen,nonacm]{acmart}

\setcitestyle{numbers,sort&compress}

\usepackage{xcolor}
\usepackage{enumitem}
\ifPDFTeX
  \usepackage[T1]{fontenc}
  \usepackage[utf8]{inputenc}
\fi
\usepackage{bbding}
\usepackage{microtype}
\usepackage{amsmath}

\usepackage{booktabs}
\usepackage{multirow}
\usepackage{needspace}

\usepackage[skins,breakable]{tcolorbox}

\usepackage{pifont}
\newcommand{\cmark}{\textcolor{green}{\ding{51}}}
\newcommand{\xmark}{\textcolor{red}{\ding{55}}}

\usepackage{colortbl}
\usepackage{makecell}
\usepackage{graphicx}
\usepackage{tikz}
\usepackage{arydshln}
\usepackage{float}      %
\usepackage{caption}    %
\usepackage{subcaption}
\usepackage{verbatim} %
\tcbuselibrary{breakable} %
\usepackage{array}
\ifPDFTeX\else
  
\fi
\usepackage{hyperref}
\usepackage[normalem]{ulem}

\usepackage{placeins}

\newcommand{\flaw}[1]{\textcolor{purple}{#1}}
\newcommand{\good}[1]{\textcolor{Green}{#1}}

\definecolor{blueText}{HTML}{5B8DB8}
\definecolor{yellowText}{HTML}{D49A5B}

\newcommand{\myblueText}[1]{\textcolor{blueText}{#1}}
\newcommand{\myyellowText}[1]{\textcolor{yellowText}{#1}}

\definecolor{myblue}{HTML}{D0E8FF}
\definecolor{myyellow}{HTML}{FBE3D2}

\usepackage{soul}

\definecolor{highlightYellow}{HTML}{FFFF00}

\newtcolorbox{calloutblock}{
    colback=white, %
    colframe=black, %
    boxrule=0.5pt, %
    arc=0pt, %
    left=1pt, %
    right=1pt, %
    top=1pt, %
    bottom=1pt, %
    breakable,
    fontupper=\footnotesize,
}

\begin{document}

\title[\textit{Standing on the Shoulders of Giants}: Enhancing LLM Agents for Automated Fact-Checking with Human Expert Skills]
{\textit{Standing on the Shoulders of Giants}: Enhancing LLM Agents for Automated Fact-Checking with Human Expert Skills}

\author{Xinwen Chen}
\authornote{Equal contribution.}
\affiliation{%
  \institution{Department of Interactive Media, Hong Kong Baptist University}
  \country{}
  }
\email{xinwwwc@gmail.com}

\author{Haorui He}
\authornotemark[1]
\affiliation{%
  \institution{Department of Interactive Media, Hong Kong Baptist University}
  \country{}
  }
\affiliation{%
  \institution{School of Computing and Data Science, The University of Hong Kong}
  \country{}
  }
\email{harryhe@connect.hku.hk}

\author{Yupeng Li}
\authornote{Corresponding author.}
\affiliation{%
  \institution{Department of Interactive Media, Hong Kong Baptist University}
  \country{}
  }
\affiliation{%
  \institution{AI and Social Good Lab, AI Media Centre, Hong Kong Baptist University}
  \country{}
  }
\email{ivanypli@gmail.com}

\begin{abstract}

Automated fact-checking (AFC) systems powered by large language model (LLM) agents still face challenges in evidence retrieval and claim verification.
Existing approaches improve their capabilities through computationally costly post-training or human-in-the-loop feedback that is difficult to scale.
To transfer professional fact-checking expertise to LLM agents efficiently, we distill publicly available guidelines published by Agence France-Presse (AFP) Fact Check into a reusable, modular Professional Fact-Checking Skill (\textbf{\textsc{PFC-Skill}}).
We then present \textsc{\textbf{SkillCheck}}, the first AFC system that gives LLM agents dynamic access to \textsc{PFC-Skill} to guide evidence retrieval and claim verification at inference time.
Extensive experiments demonstrate that \textsc{SkillCheck} improves upon skill-free DEFAME baselines across multiple LLM backbones, with gains of 2.2--4.8 pp.\ in Accuracy and 1.5--6.3 pp.\ in Macro-F1 on the AVeriTeC benchmark.
Notably, \textsc{SkillCheck} with GPT-4o-Mini exceeds DEFAME with GPT-4o by 3.4 pp.\ in Accuracy while cutting inference cost by roughly 61\%. 
The performance gains also extend to open-web retrieval.
Ablations show that incorporating the skill during evidence retrieval provides most of the gains, while incorporation into both stages yields the best overall performance.
Case studies show that \textsc{PFC-Skill} enables more targeted search and closer scrutiny of evidence.
\emph{To conclude, our work shows that LLM agents can ``stand on the shoulders of giants'' by drawing on the best practices of professional fact-checkers, substantially enhancing AFC performance without extra training or continuous human supervision.}

\end{abstract}



\maketitle

\section{Introduction}\label{sec:intro}
The rapid and widespread dissemination of misinformation presents a significant societal challenge that cannot be effectively addressed by professional human fact-checkers alone \cite{factcheck_definition,www_mcfend,guo2022survey,multimodal_survey,mdfend}. 
This limitation has motivated the development of automated fact-checking (AFC) systems to address misinformation at scale on the Web.
State-of-the-art (SOTA) AFC systems, such as DEFAME \cite{defame}, leverage large language model (LLM) agents to verify check-worthy claims.\footnote{Check-worthy claims attract public interest and can influence public behavior \cite{konstantinovskiy2021toward}.}
These systems \cite{defame,churina-etal-2024-improving,infact,xie-etal-2025-fire,hero,zheng2024evidence} typically operate in two main stages: \emph{(i) evidence retrieval (ER)}, where relevant information is iteratively collected from external sources such as the web, and \emph{(ii) claim verification (CV)}, where the system reasons over the evidence to produce a final verdict, accompanied by a justification that explains the underlying rationale \cite{guo2022survey,zeng-gao-2024-justilm,eldifrawi-etal-2024-automated}.

\begin{figure}[t]
    \centering
    \includegraphics[width=0.99\linewidth]{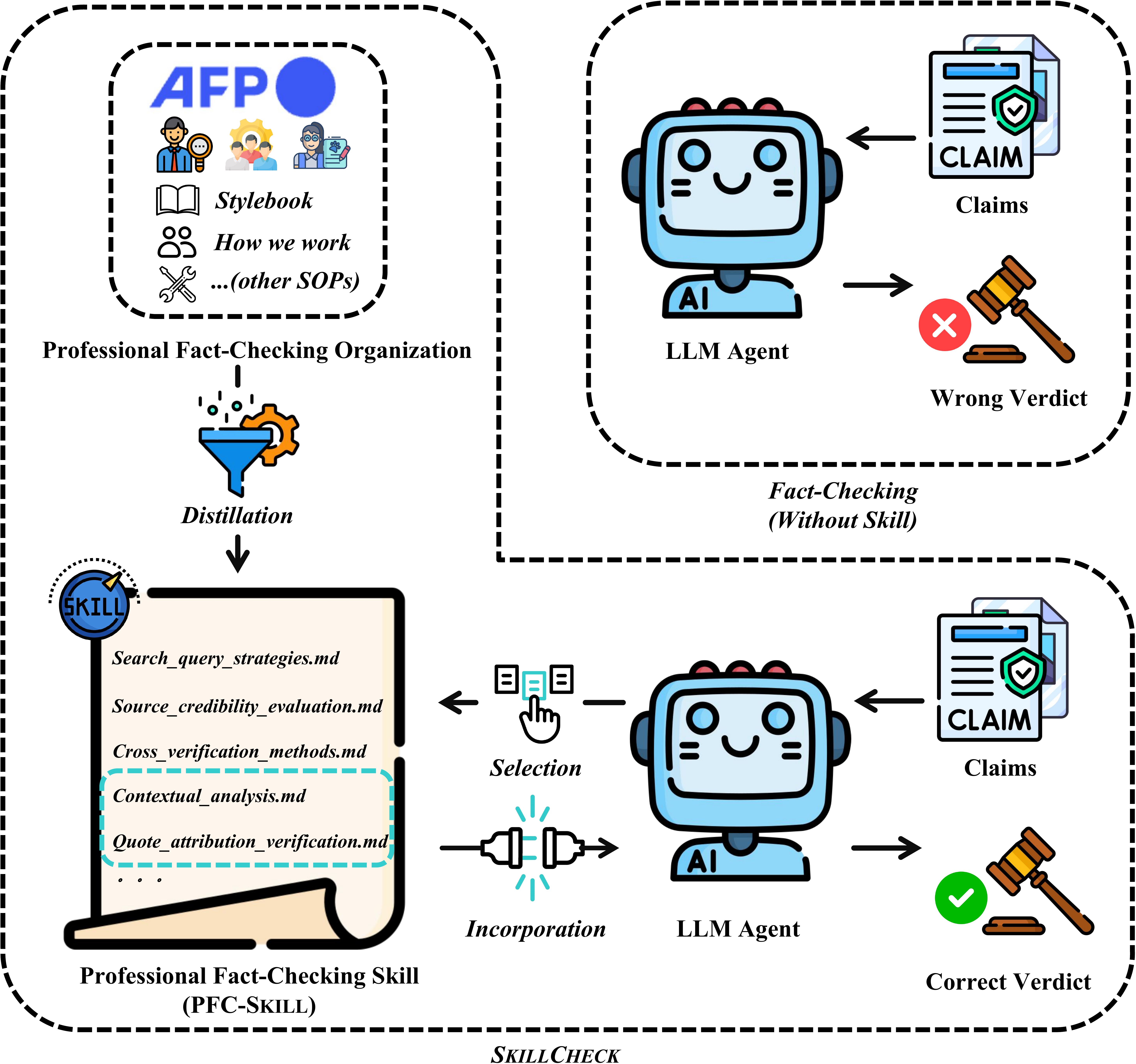}%
    \caption{Overview of LLM agents enhanced with human expert skills for automated fact-checking (AFC).}
    \label{fig:overview}
\end{figure}

\textbf{However, current LLM agents still fall short on AFC tasks.}
A significant performance gap exists between using automatically retrieved evidence and human-annotated gold evidence: with automatic retrieval, LLM agents achieve only 54.2--72.8\% Accuracy \cite{defame,infact,he2025debating}, whereas access to gold evidence boosts Accuracy to over 81.6\% \cite{he2025debating,bernardelle2026context,usmanova2026robust,momii-etal-2024-rag}.
This disparity underscores \emph{evidence retrieval} as a major bottleneck in AFC.
Furthermore, even with gold evidence, LLM agents do not achieve perfect claim verification, as their Accuracy remains below 100\%.

To improve the AFC capabilities of LLM agents, recent work has explored post-training techniques such as supervised fine-tuning (SFT) \cite{he2025debating,hero,tran-etal-2025-leaf,luo2026cheap} and reinforcement learning (RL) \cite{verri,factr1,zhang2026knowfc}.
However, these methods inevitably \textbf{incur prohibitive computational costs} \cite{greenai,bolon2024review,morrison2025environment}.
Alternatively, \citet{sahnan2026co} propose a training-free method that incorporates a human-in-the-loop mechanism at inference time to provide feedback on AFC outputs.
Yet this approach \textbf{relies heavily on manual labor and thus struggles to scale to the massive volume of misinformation}.
Consequently, closing the gap between LLMs and human experts on AFC tasks hinges on a crucial question:
\emph{How can we transfer human fact-checking expertise to LLM agents in a cost-efficient and scalable manner to enhance their AFC capabilities?}

Recent advances in agent skill use suggest a potent paradigm
for transferring human expertise to LLM agents without repeatedly involving humans in the loop~\cite{li2026skillsbench,wang2026webxskill,lu2026contractskill,deng2026harnessing}.
This approach encodes human expertise into reusable guidance (i.e., skills) that agents can consult on demand.
\textbf{However, applying this paradigm to AFC remains underexplored, raising three key questions:}
\textbf{(i)} How can we identify the best practices of professional fact-checkers and distill them into agent skills?
\textbf{(ii)} How can these skills be effectively incorporated into evidence retrieval and claim verification within AFC systems based on LLM agents?
\textbf{(iii)} What impact do these skills have on the performance and inference cost of AFC systems?

To bridge this gap, we draw on publicly available guidelines published by professional fact-checking organizations.
In particular, we focus on the guidelines of Agence France-Presse (AFP) Fact Check.\footnote{Our preliminary review of fact-checking guidelines from various organizations revealed considerable differences in procedural transparency. We selected AFP (\url{https://factcheck.afp.com/}) as our primary source due to its transparent and comprehensive guidelines, as well as its status as a certified member of the International Fact-Checking Network (IFCN), underscoring its recognized authority in fact-checking.}
These guidelines codify the standard operating procedures (SOPs) employed by AFP and detail key principles such as how to search for images, when and how to cross-check multiple independent sources, and how to verify quotes and assess factual claims.
As illustrated in Fig.~\ref{fig:overview}, 
we first distill these guidelines into a reusable, modular agent skill, termed the Professional Fact-Checking Skill (\textbf{\textsc{PFC-Skill}}).
Then, we introduce \textbf{\textsc{SkillCheck}}, the first AFC system that equips LLM agents with dynamic access to \textsc{PFC-Skill} at inference time, enabling more effective fact-checking without additional model training or human-in-the-loop feedback.

\textbf{Experiments} show that \textsc{SkillCheck} improves upon the SOTA skill-free DEFAME baselines~\cite{defame}, achieving gains of 2.2--4.8 pp.\ in Accuracy and 1.5--6.3 pp.\ in Macro-F1.
\emph{Notably, \textsc{SkillCheck} with GPT-4o-Mini surpasses DEFAME with GPT-4o by 3.4 pp.\ in Accuracy, while reducing inference cost by approximately 61\%.}
\textbf{Open-web evaluation} shows that these performance gains extend from controlled knowledge-base settings to open-web retrieval.
\textbf{Ablation studies} indicate that skill incorporation during ER provides most performance gains, while incorporation into both ER and CV yields the best AFC performance.
\textbf{Case studies} illustrate how skill guidance enables LLM agents to generate targeted queries for evidence retrieval and conduct closer scrutiny during claim verification.

Our contributions are summarized as follows:
\begin{itemize}
    \item \textbf{\textsc{PFC-Skill}.} We introduce the first agent skill for fact-checking, distilling AFP's expertise into reusable, modular guidance for evidence retrieval and claim verification.
    
    \item \textbf{\textsc{SkillCheck}.} We present the first AFC system that gives LLM agents dynamic access to \textsc{PFC-Skill} at inference time.
    
    \item \textbf{Evaluation and Analysis.} We systematically evaluate \textsc{PFC-Skill} and \textsc{SkillCheck} across different AFC stages and retrieval settings, and provide a comprehensive analysis via ablation studies, case studies, and error analysis.
\end{itemize}

\section{Related Work}\label{sec:relatedwork}

\paragraph{Automated Fact-Checking (AFC)}
As discussed above, state-of-the-art AFC systems using LLM agents \cite{guo2022survey,defame,hero,infact} still lag behind human experts in both evidence retrieval and claim verification \cite{juneja2022human,das2023state,he2025debating,fact2fiction}.
To close this gap, recent studies explore post-training via supervised fine-tuning \cite{he2025debating,hero,tran-etal-2025-leaf,luo2026cheap} or reinforcement learning \cite{verri,factr1,zhang2026knowfc}, but these methods incur prohibitive computational costs and contradict the principles of green AI \cite{greenai,patterson2021carbon,bolon2024review,morrison2025environment}. Alternatively, a concurrent study \cite{sahnan2026co} proposes a training-free human-in-the-loop framework that leverages expert feedback at inference time, yet this approach relies heavily on manual labor~\cite{lazaros2026human}.
\emph{In contrast, we introduce \textsc{SkillCheck}, a skill-based AFC system that transfers professional fact-checking know-how into a reusable, modular agent skill (\textsc{PFC-Skill}) and incorporates it into both stages without model training or human-in-the-loop feedback.}

\paragraph{Skills for LLM Agents.}
\emph{Skills} are reusable, modular procedural instructions written in natural language that encode step-by-step expertise for specific tasks and can be dynamically invoked by LLM agents at inference time \citep{anthropic2025agentskillscourse, anthropic2025skills, anthropic2025skillcreator}.
Prior work has used skills to improve 
tool use \cite{chen2026skillcraft}, 
planning \cite{wang2026skillx}, 
web navigation \cite{wang2026webxskill}, 
and task orchestration \cite{zhang2026skillflow}.
There are broadly two routes to constructing skills:
\textit{AI-crafted} skills, which are synthesized by agents from their own successful trajectories \cite{lu2026contractskill,wang2026skillx}, and \textit{human-crafted} skills, in which procedural knowledge is provided by human experts \cite{anthropic2025skills,li2026skillsbench,xu2026agent}.
Empirical evaluations report limited or negative gains from AI-crafted skills in many tasks, while human-crafted skills can provide more reliable improvements \cite{li2026skillsbench}.
These findings motivate the use of documented human expertise as the primary source of agent skills.
However, in AFC, it remains unclear how to distill human expertise into agent skills, how to incorporate them effectively into evidence retrieval and claim verification, and how they affect LLM agents' AFC performance.
We investigate these questions by distilling AFP's publicly available guidelines into \textsc{PFC-Skill} and evaluating its incorporation into both stages through \textsc{SkillCheck}.

\begin{figure*}[t]
\centering
\includegraphics[width=0.76\linewidth]
{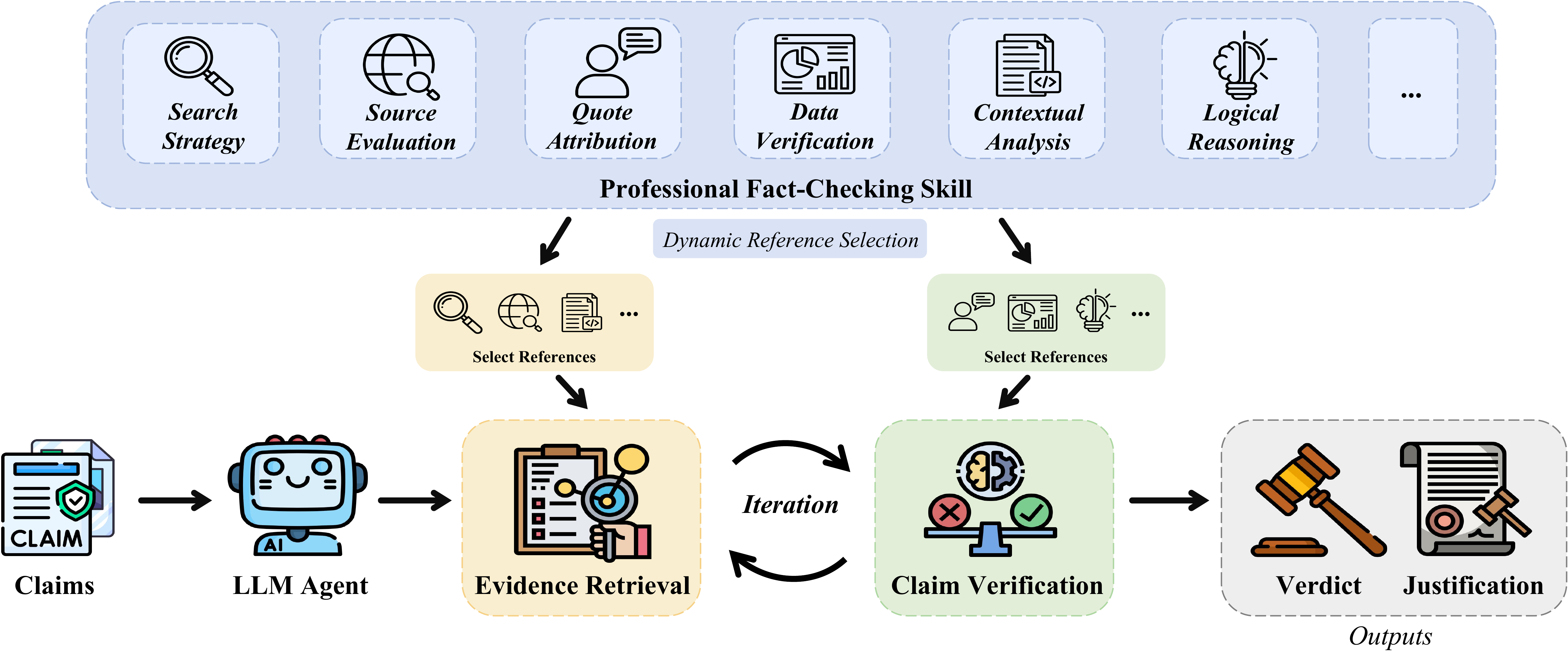}
\caption{\textsc{PFC-Skill} is incorporated into the evidence retrieval and claim verification stages of an AFC system.}
\label{fig:pip}
\end{figure*}

\section{The \textsc{SkillCheck} System}
\label{sec:skillcheck}
In this section, we introduce the \textsc{SkillCheck} system by detailing its two core components: \textbf{skill creation}, which builds the unified \textsc{PFC-Skill} from the AFP guidelines, and \textbf{skill incorporation}, which provides agents with targeted guidance during fact-checking.

\subsection{Skill Creation}
\label{sec:skill_construction}
Human fact-checking experts document their best practices for evidence retrieval and claim verification on their websites to ensure transparency (e.g., in the AFP guidelines).
To make this knowledge usable by LLM agents, we extract executable procedures and organize them into a Professional Fact-Checking Skill (\textsc{PFC-Skill}).

\paragraph{Skill Distillation.}
We use two key AFP documents as sources: the \textit{AFP Fact-Checking Stylebook}~\cite{afp_stylebook} and \textit{How We Work}~\cite{afp_how_we_work}. 
These documents detail professional practices for gathering evidence, evaluating sources, cross-checking information, and verifying claims.
Then, we employ Anthropic's Skill Creator framework~\cite{anthropic2025skillcreator} together with Claude Opus 4.7 to formalize the procedures in the AFP guidelines into executable skill instructions, without introducing any additional fact-checking rules.
Because the guidelines also cover editorial policies and tasks specific to human fact-checkers, we restrict skill creation to procedures directly relevant to evidence retrieval and claim verification.
By default, we distill procedures for both stages into a unified \textsc{PFC-Skill}.
As an alternative, we also consider separating the stages and distilling them individually; however, as discussed in Sec.~\ref{sec:rq2}, this stage-specific approach does not yield consistent improvements over the unified \textsc{PFC-Skill}.

\begin{figure}[t]
\centering
\includegraphics[width=0.96\linewidth]
{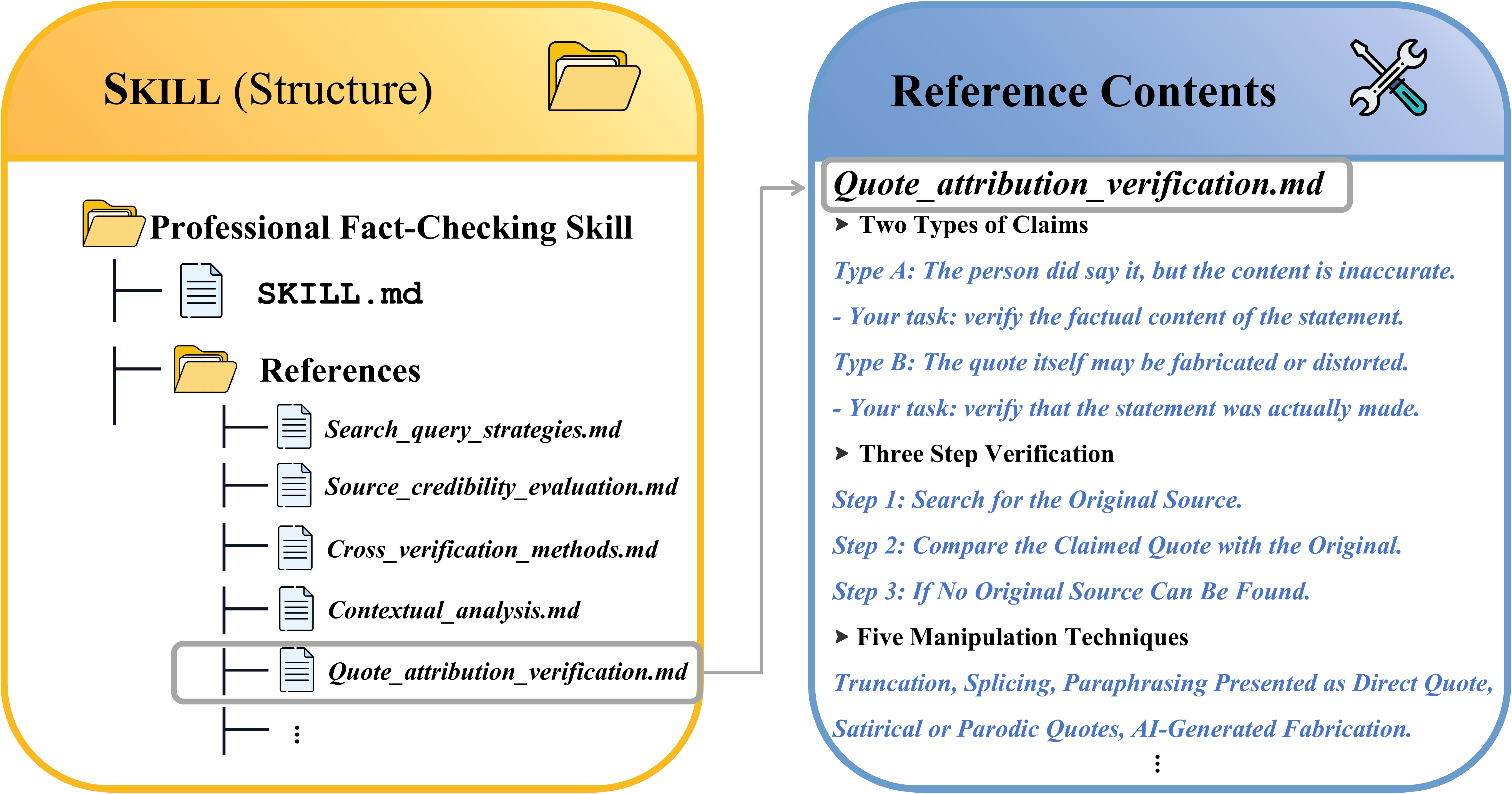}
\caption{Structure of \textsc{PFC-Skill}.}
\label{fig:skill_construct}
\end{figure}

\paragraph{Skill Organization.}
To make the distilled procedures accessible on demand, we structure \textsc{PFC-Skill} into two main components following prior work~\cite{anthropic2025skills,anthropic2025skillcreator}, as illustrated in Fig.~\ref{fig:skill_construct}: high-level skill-usage instructions (\texttt{SKILL.md}) and a set of reference files.\footnote{Appendix~\ref{app:skill_details} provides a complete description of all reference files in \textsc{PFC-Skill}.}
Formally, we represent \textsc{PFC-Skill} as
$\mathcal{S} = (I, R)$, where $I$ denotes the instructions in \texttt{SKILL.md} and $R$ denotes the collection of reference files.
The instruction file guides agents on when and how to apply the skill, as well as how to utilize the relevant reference files during the process.
Each reference file $r_i \in R$ outlines the procedure for a particular fact-checking subtask, such as claim decomposition, source credibility assessment, or cross-verification.
For example, the cross-verification reference file provides detailed steps for evaluating source independence, comparing evidence across sources, and resolving conflicting information.
This organization enables agents to efficiently access procedural guidance during fact-checking.

\subsection{Skill Incorporation}
\label{sec:skill_incorporation}
To equip LLM agents with access to \textsc{PFC-Skill} during fact-checking, we build \textsc{SkillCheck} on DEFAME~\cite{defame}, a state-of-the-art AFC system.
DEFAME follows the standard ReAct-style~\cite{react} agentic paradigm of iteratively reasoning and selecting actions, making it a natural foundation for incorporating procedural skills.
Given a claim \(c\), the agent plans search queries, examines retrieved evidence, and ultimately produces a verdict together with a justification based on the accumulated fact-checking record.
Formally, the baseline process can be written as
$
y = F(c) = V\!\left(c, E(c)\right),
$
where \(F(\cdot)\) denotes the AFC system, \(E(\cdot)\) denotes evidence retrieval, \(V(\cdot)\) denotes claim verification, and \(y\) is the predicted veracity label.

We incorporate \textsc{PFC-Skill} into this AFC system by conditioning both evidence retrieval and claim verification on the skill:
$
y = F(c; \mathcal{S}) = V\!\left(c, E(c; \mathcal{S}), \mathcal{S}\right),
$
where \(\mathcal{S} = (I, R)\) denotes \textsc{PFC-Skill}.
However, naively injecting the entire \textsc{PFC-Skill} into the context presents practical challenges, as excessively long contexts can hinder the retrieval of relevant information when literal matching cues are absent~\cite{modarressi2025nolima}.
To retain focused procedural guidance, \textsc{SkillCheck} dynamically selects reference files from \(R\), as illustrated in Fig.~\ref{fig:pip}.\footnote{Appendix~\ref{app:prompt} provides an example prompt template for skill incorporation.}
Concretely, at each step $t$, the acting LLM agent reads the instructions $I$ in \texttt{SKILL.md} alongside the reference catalog, and selects a relevant subset of reference files conditioned on the claim, the current stage, and the accumulated record of actions and observations.
The agent outputs a subset \(R_t \subseteq R\), or \texttt{None} when no reference files are relevant.
The contents of the selected reference files are loaded into the agent's context to guide the corresponding ER or CV step.
In this way, \textsc{SkillCheck} enables LLM agents to leverage professional fact-checking expertise on demand, without real-time human supervision or additional model training.

\section{Experiments}\label{sec:experiments}
In this section, we conduct experiments to evaluate \textsc{SkillCheck}.

\subsection{Experimental Setup}\label{sec:experiment_setup}
\paragraph{\textbf{Datasets.}}
We use \textbf{AVeriTeC}~\cite{averitec} as our primary benchmark. It comprises English textual claims collected from 50 real-world fact-checking organizations and annotated with four veracity labels: \textsc{Refuted}, \textsc{Supported}, \textsc{NEI}, and \textsc{Conflicting/Cherry-picking}.\footnote{See Appendix~\ref{app:label1} for detailed definitions of the AVeriTeC labels.}
Unlike earlier datasets such as FEVER~\cite{fever}, AVeriTeC mitigates evidence leakage and insufficiency~\cite{averitec,averitec_challenge}, rendering it a more reliable benchmark for assessing automated fact-checking (AFC) systems under realistic conditions.
We use the 500-claim development split of AVeriTeC.
We remove duplicate claims based on claim text, resulting in 491 unique claims for evaluation.
Each claim is accompanied by a knowledge base (KB) that contains both supporting evidence and a large number of distractor documents.
This setup simulates open-web retrieval while ensuring reproducibility \cite{averitec_challenge}.
Following prior work~\cite{infact,hero,defame,he2025debating}, for each query, we retrieve the top five evidence items from the KB via semantic search with \texttt{gte-base-en-v1.5}~\cite{zhang2024mgte}.\footnote{Prior studies show that LLM-agent-based AFC systems perform robustly across different retrievers~\cite{liu-etal-2026-compete,he2025debating}.}
To further assess whether the performance gains generalize to an open-web setting, 
we evaluate \textsc{SkillCheck} on \textbf{ClaimReview2024+}~\cite{defame}, a collection of 300 English claims drawn from the ClaimReview project.\footnote{\url{https://www.claimreviewproject.com/}}
Each claim is annotated with one of four veracity labels: \textsc{Refuted}, \textsc{Supported}, \textsc{Misleading}, or \textsc{NEI}.\footnote{See Appendix~\ref{app:label2} for detailed definitions of the ClaimReview2024+$_{\text{date}}$ labels.}
These claims were drawn from fact-checking articles published from November 2023 onward, after GPT-4o's October 2023 knowledge cutoff, reducing the risk of knowledge contamination~\cite{defame,rethink_mafc_eval_2026}.
To avoid temporal leakage in evidence, we retain only claims with explicit release dates, resulting in a filtered subset of 199 claims, denoted \textbf{ClaimReview2024+$_{\text{date}}$}. For each claim, we restrict evidence to documents published before the claim's release date. We use date-range filters (e.g., \texttt{before:YYYY-MM-DD}) during search.
On this dataset, LLM agents interact with the open web via Google Web Search, Image Search, Reverse Image Search, and geolocation tools through Serper\footnote{\url{https://serper.dev/}} and Google Vision.\footnote{\url{https://cloud.google.com/vision}}

\paragraph{\textbf{Baselines.}} \label{sec:baselines}
We compare \textsc{SkillCheck} against two categories of baselines: \textbf{post-training methods} and \textbf{training-free methods}.

\textbf{Post-training methods} enhance LLM agents' fact-checking performance via additional training. 
For example, \textit{HerO}~\cite{hero} applies supervised fine-tuning (SFT) on the AVeriTeC training split.
Similarly, \textit{DebateCV}~\cite{he2025debating} employs opposing debater agents and fine-tunes a moderator on synthetic debate transcripts to generate verdicts and justifications.
Both methods rely on fine-tuned LLM agents;
we report results for their \textit{Llama-3.1-8B} implementations as our baselines.
Results for both baselines are taken from \cite{he2025debating}.

\textbf{Training-free methods} perform fact-checking without any fine-tuning.
\textit{Zero-shot} directly prompts an LLM to predict a veracity label from the claim and evidence. 
\textit{Chain-of-Thought (CoT)} prompts the LLM to reason step by step before producing a verdict.
Results for both \textit{Zero-shot} and \textit{CoT} are taken from \citep{defame}. 
\textit{DebateCV (Zero-shot)}~\cite{he2025debating} follows the same workflow as DebateCV, but without any moderator fine-tuning; its results are also taken from \citep{he2025debating}. 
Finally, \textit{DEFAME}~\cite{defame}, the SOTA training-free AFC system and our main skill-free baseline, iteratively plans search actions, retrieves evidence, and updates its fact-checking record before generating a verdict and justification.
We use DEFAME's official codebase\footnote{\url{https://github.com/multimodal-ai-lab/DEFAME}}.

\paragraph{\textbf{Ablations.}} \label{sec:ablations}
To assess our design choices, we examine how \textsc{PFC-Skill} is created and where it is incorporated into AFC.
For \textbf{skill creation}, we first compare \textsc{SkillCheck} with skill-free DEFAME (\textit{w/o Skill}) to assess the benefit of \textsc{PFC-Skill}. We then compare \textsc{SkillCheck}, which uses distilled \textsc{PFC-Skill}, with the \textit{w/o Distillation} variant, which provides raw AFP guidelines during evidence retrieval (ER) and claim verification (CV), to assess the added value of distillation.
Since ER and CV involve distinct procedures, we also compare \textsc{SkillCheck} with stage-specific skills (\textit{w/ Stage-Specific Skill}), which provide separate guidance for ER and CV.
For \textbf{skill incorporation}, we isolate the effects of guidance at each stage by comparing three settings: skill-free DEFAME (\textit{w/o ER \& CV}, equivalent to \textit{w/o Skill}), guidance during ER only (\textit{w/o CV}), and guidance during both ER and CV (\textsc{SkillCheck}).

\paragraph{\textbf{LLM Backbones.}}
Our experiments cover \textbf{closed-source} and \textbf{open-weight} LLM backbones.
The closed-source models are \textit{GPT-4o-2024-08-06} (hereafter \textit{GPT-4o}) and \textit{GPT-4o-Mini-2024-07-18} (\textit{GPT-4o-Mini}).
We also evaluate the open-weight \textit{DeepSeek-V4-Flash-0731} (hereafter \textit{DeepSeek-V4-Flash}) and \textit{Qwen3.5-35B-A3B}.
We use a temperature of 0.01 to reduce sampling variability.

\paragraph{\textbf{Metrics.}}
We use two metrics: (1) \textbf{Accuracy (Acc.)}, defined as the proportion of correctly classified claims; and (2) \textbf{Macro-F1}, the unweighted mean F1 score across all classes.
Results are averaged over at least three repeated runs.
For statistical rigor, we report 95\% bootstrap confidence intervals (CIs) for the mean scores, computed using 10,000 resamples of matched claims.

\begin{table*}[tp]
\centering
\normalsize
\caption{
AFC results on AVeriTeC.
\good{Green} superscripts indicate absolute improvements, in pp., over the corresponding skill-free DEFAME setting.
For DEFAME and \textsc{SkillCheck}, uncertainty is reported as 95\% bootstrap confidence intervals on matched claims.
$\ast$ indicates statistically significant improvements ($p<0.05$, paired bootstrap test).
}
\label{tab:main_results}

\begin{tabular}{llcccc}
\toprule

\multirow{2}{*}{\textbf{Method}}
&
\multirow{2}{*}{\textbf{LLM}}
&
\multicolumn{2}{c}{\textbf{Accuracy}}
&
\multicolumn{2}{c}{\textbf{Macro-F1}}
\\

\cmidrule(lr){3-4}
\cmidrule(lr){5-6}

&
&
\textbf{Score (\%)}
&
\textbf{95\% CI}
&
\textbf{Score (\%)}
&
\textbf{95\% CI}
\\

\midrule

\multicolumn{6}{c}{\textbf{Post-training methods}}
\\
\midrule

DebateCV
& Llama-3.1-8B
& 72.8
& -
& -
& -
\\

HerO
& Llama-3.1-8B
& 70.2
& -
& -
& -
\\

\midrule

\multicolumn{6}{c}{\textbf{Training-free methods}}
\\
\midrule

Zero-shot
& GPT-4o
& 62.1
& -
& -
& -
\\

CoT
& GPT-4o
& 61.0
& -
& -
& -
\\

DebateCV (Zero-shot)
& GPT-4o
& 65.4
& -
& -
& -
\\

\hdashline

\multirow{4}{*}{DEFAME}
& GPT-4o
& 70.8
& [66.8, 74.8]
& 43.9
& [39.8, 48.3]
\\

& GPT-4o-Mini
& 69.4
& [65.4, 73.4]
& 41.7
& [37.6, 45.9]
\\

& DeepSeek-V4-Flash
& 73.9
& [69.9, 77.7]
& 46.2
& [41.5, 51.1]
\\

& Qwen3.5-35B-A3B
& 66.9
& [62.7, 71.1]
& 42.7
& [38.5, 46.9]
\\

\hdashline

\multirow{4}{*}{\textsc{SkillCheck}}
& GPT-4o
& 74.8\rlap{%
\textsuperscript{\textcolor{ForestGreen}{\scriptsize +4.0*}}}
& [71.0, 78.6]
& 50.2\rlap{%
\textsuperscript{\textcolor{ForestGreen}{\scriptsize +6.3*}}}
& [45.1, 55.3]
\\

& GPT-4o-Mini
& 74.2\rlap{%
\textsuperscript{\textcolor{ForestGreen}{\scriptsize +4.8*}}}
& [70.4, 78.0]
& 47.3\rlap{%
\textsuperscript{\textcolor{ForestGreen}{\scriptsize +5.6}}}
& [42.9, 51.9]
\\

& DeepSeek-V4-Flash
& 76.9\rlap{%
\textsuperscript{\textcolor{ForestGreen}{\scriptsize +3.0*}}}
& [73.1, 80.5]
& 47.7\rlap{%
\textsuperscript{\textcolor{ForestGreen}{\scriptsize +1.5}}}
& [43.0, 52.8]
\\

& Qwen3.5-35B-A3B
& 69.1\rlap{%
\textsuperscript{\textcolor{ForestGreen}{\scriptsize +2.2}}}
& [65.1, 73.2]
& 46.7\rlap{%
\textsuperscript{\textcolor{ForestGreen}{\scriptsize +4.0}}}
& [41.9, 51.4]
\\

\bottomrule
\end{tabular}

\end{table*}

\subsection{AFC Performance and Inference Efficiency}\label{sec:main_results}

Table~\ref{tab:main_results} compares \textsc{SkillCheck} against baselines on AVeriTeC.
The results show that \textsc{SkillCheck} substantially outperforms all training-free baselines.
With GPT-4o, \textsc{SkillCheck} achieves 74.8\% Accuracy, exceeding the Zero-shot, CoT, and DebateCV (Zero-shot) baselines by 12.7, 13.8, and 9.4 pp., respectively (62.1\%, 61.0\%, and 65.4\%).
For comparisons with DEFAME, we employ paired bootstrap tests to assess statistical significance. \textbf{\textsc{SkillCheck} significantly outperforms ($p<0.05$) the state-of-the-art training-free baseline DEFAME in Accuracy across most of the tested LLM backbones.}
Specifically, Accuracy increases by 4.0, 4.8, 3.0, and 2.2 pp.\ for GPT-4o, GPT-4o-Mini, DeepSeek-V4-Flash, and Qwen3.5-35B-A3B, respectively.
Additionally, \textsc{SkillCheck} with DeepSeek-V4-Flash outperforms post-training methods such as DebateCV (72.8\% Accuracy) and HerO (70.2\% Accuracy), achieving 76.9\% Accuracy.

\textbf{We also observe that \textsc{PFC-Skill} gains vary across LLM backbones, indicating differences in how effectively models apply agent skills.}
This parallels findings in tool learning, where LLMs vary in their ability to use tools effectively~\cite{chen2026skillcraft}.
These results motivate \emph{skill learning} as a future direction for next-generation LLMs: improving their ability to apply skills during task execution.

\begin{table*}[tp]
\centering
\caption{Average token usage, LLM inference cost, and latency per claim. Token counts are reported in thousands (K). Input, output, and total cost are rounded independently.}
\label{tab:cost}
\begin{tabular}{llrrrrrr}
\toprule
\multirow{2}{*}{\textbf{Backbone}} & \multirow{2}{*}{\textbf{System}}
& \multicolumn{2}{c}{\textbf{Input}} & \multicolumn{2}{c}{\textbf{Output}}
& \textbf{Total Cost} & \textbf{Latency} \\
\cmidrule(lr){3-4}\cmidrule(lr){5-6}
& & Tokens (K) & Cost (\$) & Tokens (K) & Cost (\$) & (\$) & (s) \\
\midrule
\multirow{2}{*}{GPT-4o} & DEFAME & 2.97 & 0.0074 & 0.37 & 0.0037 & 0.0111 & 43.11 \\
& \textsc{SkillCheck} & 26.41 & 0.0660 & 0.43 & 0.0043 & 0.0703 & 51.33 \\
\midrule
\multirow{2}{*}{GPT-4o-Mini} & DEFAME & 2.89 & 0.0004 & 0.39 & 0.0002 & 0.0007 & 43.34 \\
& \textsc{SkillCheck} & 26.75 & 0.0040 & 0.55 & 0.0003 & 0.0043 & 53.41 \\
\bottomrule
\end{tabular}
\end{table*}

\begin{figure}[!t]
\centering
\includegraphics[width=\linewidth]{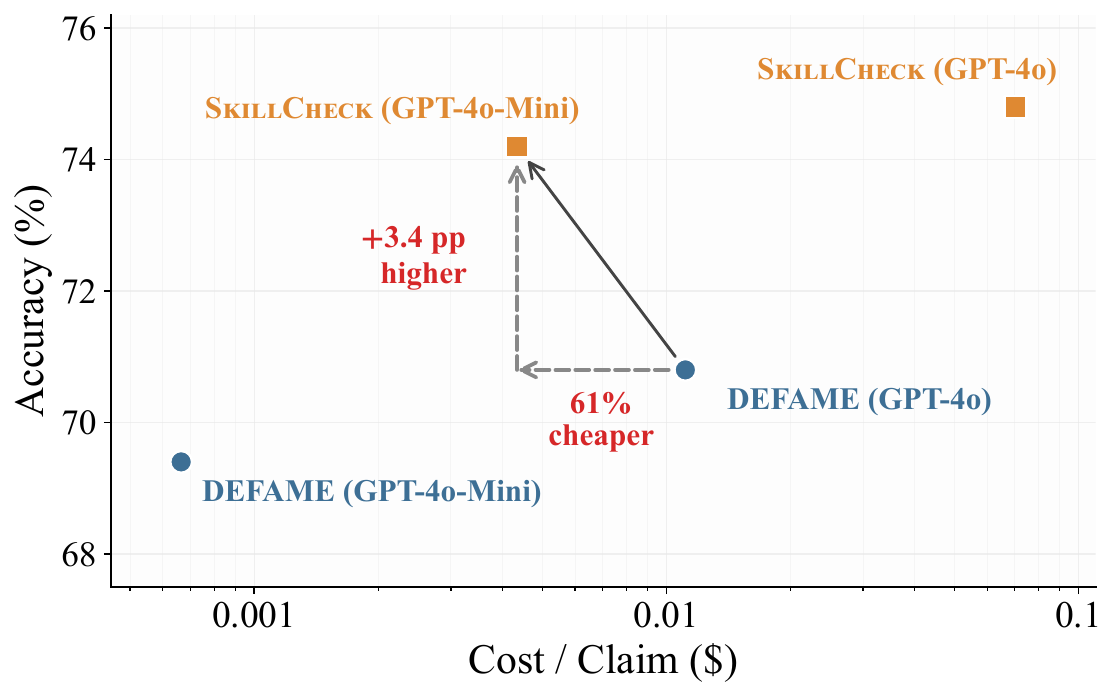}
\caption{
Comparison of Accuracy and inference cost per claim between \textsc{SkillCheck} and \textit{w/o Skill} (i.e., DEFAME).
}
\label{fig:cost}

\end{figure}

Beyond AFC performance, we further assess the inference cost of \textsc{SkillCheck}.
Following prior studies~\cite{defame,he2025debating}, we calculate the cost of each LLM call by multiplying its input and output token counts by the corresponding model's prices.\footnote{For these cost estimates, the input/output rates per million tokens are \$2.50/\$10.00 for GPT-4o and \$0.15/\$0.60 for GPT-4o-Mini.}
We sum these costs over all agent calls and average over evaluated claims.
Table~\ref{tab:cost} reports average token usage, LLM inference cost, and latency per claim on AVeriTeC, allowing comparisons within the same backbone and across model choices.
For the same backbone, incorporating \textsc{PFC-Skill} increases inference overhead. For example, with GPT-4o, the average cost per claim rises from \$0.0111 to \$0.0703, while latency grows from 43.11\,s to 51.33\,s.
Most of the added cost comes from input token processing.
Despite this overhead, \textbf{\textsc{SkillCheck} with GPT-4o-Mini achieves higher AFC performance at a lower LLM inference cost than DEFAME with GPT-4o}. As shown in Table~\ref{tab:main_results} and Fig.~\ref{fig:cost}, Accuracy improves from 70.8\% to 74.2\% and Macro-F1 from 43.9\% to 47.3\%, yielding gains of 3.4 pp.\ on both metrics. Concurrently, the average LLM inference cost per claim drops from \$0.0111 to \$0.0043, a reduction of approximately 61\%.
\textbf{This comparison highlights \textsc{SkillCheck} as a practical and cost-effective alternative to the existing SOTA skill-free AFC system, delivering higher AFC performance at a lower LLM inference cost.}

\begin{figure*}[tp]
\centering

\hspace*{0.04\textwidth}
\begin{subfigure}{0.45\textwidth}
    \centering
    \includegraphics[width=\linewidth]{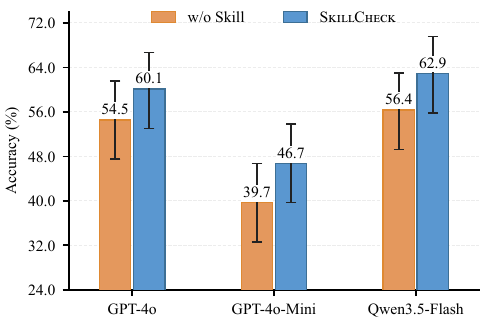}
    \caption{Accuracy}
    \label{fig:claimreview-results-acc}
\end{subfigure}
\hfill
\begin{subfigure}{0.45\textwidth}
    \centering
    \includegraphics[width=\linewidth]{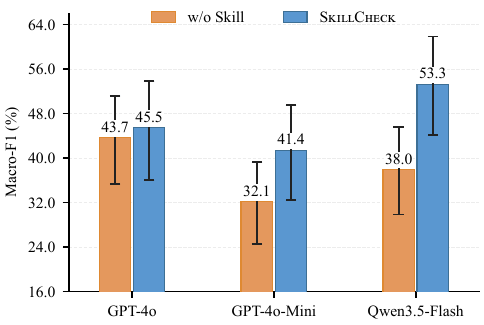}
    \caption{Macro-F1}
    \label{fig:claimreview-results-f1}
\end{subfigure}
\hspace*{0.04\textwidth}

\caption{
AFC results on ClaimReview2024+$_{\text{date}}$.
}
\label{fig:claimreview_results}
\end{figure*}

\subsection{Open-Web Evaluation}\label{sec:rq4}
The open web exposes agents to a broader and less controlled range of evidence sources.
We therefore examine whether the performance gains observed above extend to open-web evidence retrieval.

Following the experimental setup in Sec.~\ref{sec:experiment_setup}, we evaluate \textsc{SkillCheck} on ClaimReview2024+$_{\text{date}}$ using GPT-4o, GPT-4o-Mini, and Qwen3.5-35B-A3B.
Some claims in ClaimReview2024+$_{\text{date}}$ contain both text and images.
We therefore exclude DeepSeek-V4-Flash because it does not support these multimodal inputs.
As shown in Fig.~\ref{fig:claimreview_results}, \textbf{\textsc{SkillCheck}, which incorporates \textsc{PFC-Skill}, improves AFC performance with open-web evidence retrieval across all three evaluated backbones.}
Accuracy increases from 54.5\% to 60.1\% for GPT-4o, from 39.7\% to 46.7\% for GPT-4o-Mini, and from 56.4\% to 62.9\% for Qwen3.5-35B-A3B.
Macro-F1 also improves for all three backbones, by 1.8, 9.3, and 15.3 pp., respectively.
\emph{These results support incorporating \textsc{PFC-Skill} into LLM agents for AFC under both open-web and fixed-KB evidence retrieval.}

\begin{figure*}[!t]
\centering

\hspace*{0.04\textwidth}
\begin{subfigure}{0.45\textwidth}
    \centering
    \includegraphics[width=\linewidth]{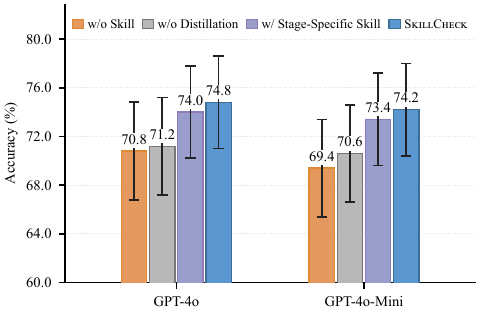}
    \caption*{Accuracy}
\end{subfigure}
\hfill
\begin{subfigure}{0.45\textwidth}
    \centering
    \includegraphics[width=\linewidth]{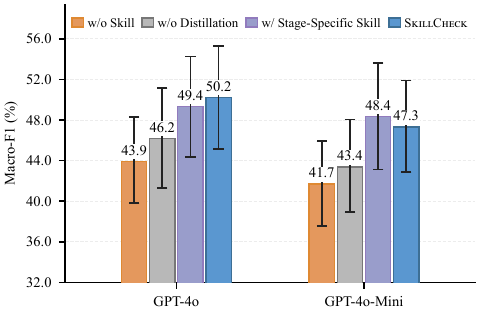}
    \caption*{Macro-F1}
\end{subfigure}
\hspace*{0.04\textwidth}

\makebox[\textwidth]{\textbf{(a) Skill Creation}}
\par

\hspace*{0.04\textwidth}
\begin{subfigure}{0.45\textwidth}
    \centering
    \includegraphics[width=\linewidth]{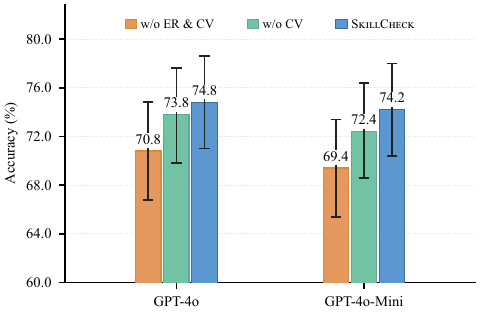}
    \caption*{Accuracy}
\end{subfigure}
\hfill
\begin{subfigure}{0.45\textwidth}
    \centering
    \includegraphics[width=\linewidth]{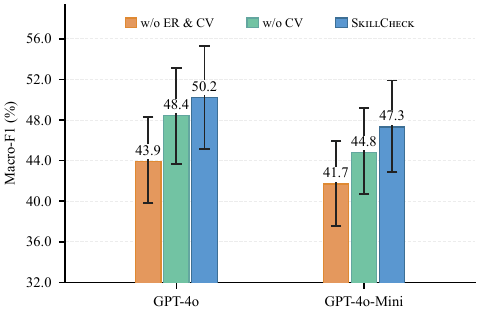}
    \caption*{Macro-F1}
\end{subfigure}
\hspace*{0.04\textwidth}

\makebox[\textwidth]{\textbf{(b) Skill Incorporation}}

\caption{
Ablation results.
(a) compares skill-free DEFAME (\textit{w/o Skill}), raw AFP guidelines (\textit{w/o Distillation}), stage-specific skills (\textit{w/ Stage-Specific Skill}), and the unified \textsc{PFC-Skill} (\textsc{SkillCheck}).
(b) compares skill-free DEFAME (\textit{w/o ER \& CV}), skill incorporation during ER alone (\textit{w/o CV}), and incorporation into both ER and CV (\textsc{SkillCheck}).
}
\label{fig:stage_analysis}

\end{figure*}

\subsection{Ablation Studies}\label{sec:rq2}
To explain the performance gains, we compare the ablation variants defined in Sec.~\ref{sec:ablations} on AVeriTeC using GPT-4o and GPT-4o-Mini.

\paragraph{\textbf{Skill Creation.}}
As shown in Fig.~\ref{fig:stage_analysis}(a), \textbf{\textsc{PFC-Skill} is more effective than directly using the raw AFP guidelines.}
The \textit{w/o Distillation} variant improves Accuracy over \textit{w/o Skill} by only 0.4 and 1.2 pp.\ for GPT-4o and GPT-4o-Mini, respectively.
\textsc{PFC-Skill} adds a further 3.6 pp.\ in Accuracy for both backbones and 4.0 and 3.9 pp.\ in Macro-F1, respectively.
\emph{These results suggest that access to professional knowledge alone provides limited benefits.}
This may be because the raw guidelines combine broad principles with procedures for human fact-checkers.
Distilling them into \textsc{PFC-Skill} makes the relevant knowledge actionable for LLM agents by specifying procedures for evidence retrieval and claim verification.
Among the distilled variants, \textbf{\textit{w/ Stage-Specific Skill} provides no consistent advantage over \textsc{SkillCheck} with the unified \textsc{PFC-Skill}}.
For GPT-4o, \textit{w/ Stage-Specific Skill} reduces both Accuracy and Macro-F1 by 0.8 pp.\ compared with \textsc{SkillCheck}.
For GPT-4o-Mini, it reduces Accuracy by 0.8 pp.\ but increases Macro-F1 by 1.1 pp.
\emph{We therefore use a single skill package for both stages, avoiding the need to construct and maintain separate stage-specific skills.}

\paragraph{\textbf{Skill Incorporation.}}
As shown in Fig.~\ref{fig:stage_analysis}(b), \textbf{the \textit{w/o CV} variant accounts for most of the performance gains, while extending skill use to CV provides additional improvements}.
The \textit{w/o CV} variant improves Accuracy and Macro-F1 by 3.0 and 4.5 pp.\ for GPT-4o, and by 3.0 and 3.1 pp.\ for GPT-4o-Mini.
These gains account for approximately 55--75\% of the improvements achieved by \textsc{SkillCheck}.
Adding CV guidance in \textsc{SkillCheck} further improves Accuracy and Macro-F1 by 1.0 and 1.8 pp.\ for GPT-4o, 
and by 1.8 and 2.5 pp.\ for GPT-4o-Mini.
\emph{These results indicate that our \textsc{PFC-Skill} offers complementary benefits at different stages, motivating skill incorporation at both fact-checking stages.}

\begin{table*}[tp]
\centering
\small

\newcommand{\caseQueryIcon}{%
\tikz[x=1.35ex,y=1.35ex,baseline=-0.2ex,draw=blueText,line width=0.45pt]{%
\draw (0.65,0.65) circle (0.55);
\draw (1.04,0.26)--(1.55,-0.25);}}

\newcommand{\caseEvidenceIcon}{%
\tikz[x=1.35ex,y=1.35ex,baseline=-0.2ex,draw=blueText,line width=0.45pt]{%
\draw (0,0)--(0,1.5)--(0.85,1.5)--(1.25,1.1)--(1.25,0)--cycle;
\draw (0.85,1.5)--(0.85,1.1)--(1.25,1.1);
\draw (0.25,0.8)--(1,0.8)
      (0.25,0.5)--(1,0.5)
      (0.25,0.2)--(0.8,0.2);}}

\newcommand{\caseJudgmentIcon}{%
\tikz[x=1.35ex,y=1.35ex,baseline=-0.2ex,draw=blueText,line width=0.45pt]{%
\draw (0.8,0)--(0.8,1.5)
      (0.25,0)--(1.35,0)
      (0.05,1.15)--(1.55,1.15);
\draw (0.25,1.15)--(0,0.6)--(0.5,0.6)--cycle;
\draw (1.35,1.15)--(1.1,0.6)--(1.6,0.6)--cycle;}}

\caption{
Case studies illustrating how the skill alters fact-checking behaviors. Selected search queries are shown; evidence and judgments are abridged.
Effective behaviors, useful skills, and correct judgments are highlighted in \good{green},
while problematic behaviors and incorrect judgments are marked in \flaw{purple}.
Positive cases are highlighted in \myblueText{blue},
while failures are in \myyellowText{yellow}.
}
\label{tab:rq2_cases1}

\resizebox{0.95\textwidth}{!}{%
\begin{tabular}{@{}
>{\raggedright\arraybackslash}p{1.7cm}
>{\raggedright\arraybackslash}p{\dimexpr0.41\textwidth-0.85cm-4\tabcolsep\relax}
>{\raggedright\arraybackslash}p{\dimexpr0.41\textwidth-0.85cm-4\tabcolsep\relax}
>{\centering\arraybackslash}p{0.26\textwidth}
@{}}

\toprule
&
\multicolumn{1}{c}{\textbf{DEFAME}}
&
\multicolumn{2}{c}{\textbf{\textsc{SkillCheck}}}
\\
\cmidrule(lr){2-2}
\cmidrule(lr){3-4}
&
\multicolumn{1}{c}{\textbf{Fact-Checking Record}}
&
\multicolumn{1}{c}{\textbf{Fact-Checking Record}}
&
\multicolumn{1}{c}{\textbf{Selected Skill}}
\\
\midrule


\multicolumn{4}{@{}>{\columncolor{myblue}}p{\dimexpr1.05\textwidth-2\tabcolsep+0.3cm\relax}@{}}{%
\textbf{Claim 1, \textsc{Refuted}:}
US President Trump made vaccines voluntary, not mandatory, in the US.
}\\

\addlinespace[2pt]

\caseQueryIcon\hspace{0.4em}\textbf{Query}
&
\begin{minipage}[t]{\linewidth}
\raggedright
\par\hangindent=1.6em\hangafter=1\noindent
\makebox[1.6em][l]{\textcolor{blueText}{1.}}
``US President Trump vaccines voluntary~mandatory~September~2020''\par
\end{minipage}
&
\begin{minipage}[t]{\linewidth}
\raggedright
\par\hangindent=1.6em\hangafter=1\noindent
\makebox[1.6em][l]{\textcolor{blueText}{1.}}
``US President Trump vaccines voluntary not~mandatory~September~2020'';

\par\hangindent=1.6em\hangafter=1\noindent
\makebox[1.6em][l]{\textcolor{blueText}{2.}}
``Trump \good{\textit{\underline{statement}}} vaccines voluntary~mandatory~September~2020'';

\par\hangindent=1.6em\hangafter=1\noindent
\makebox[1.6em][l]{\textcolor{blueText}{3.}}
``Trump executive order vaccines voluntary~mandatory~September~2020'';

\par\hangindent=1.6em\hangafter=1\noindent
\makebox[1.6em][l]{\textcolor{blueText}{4.}}
``\good{\textit{\underline{US federal}}} vaccine~\good{policy}~September~2020''.\par
\end{minipage}
&
\multirow{3}{*}{%
\includegraphics[width=0.25\textwidth]{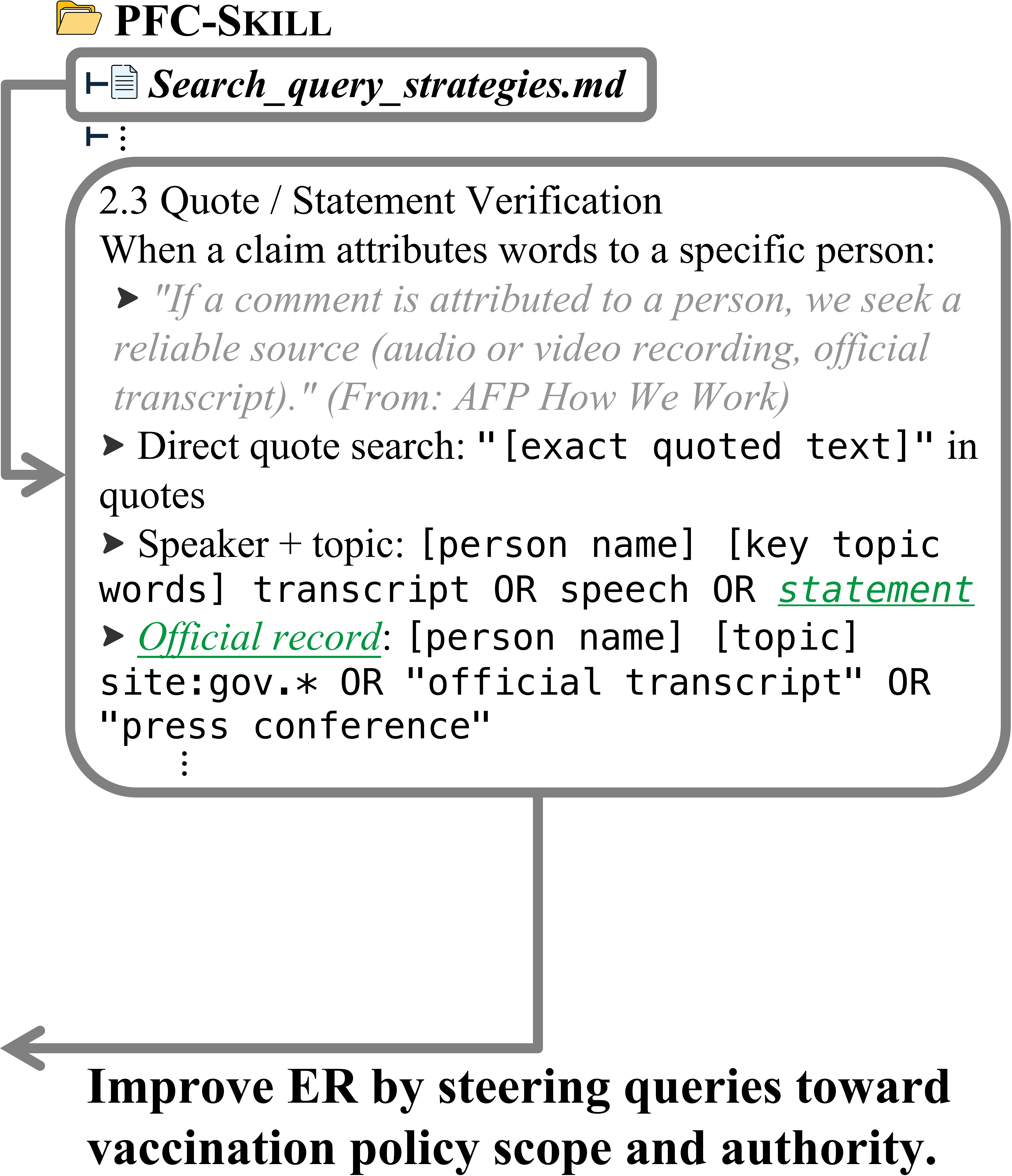}
}
\\

\addlinespace[2pt]

\caseEvidenceIcon\hspace{0.4em}\textbf{Evidence}
&
Trump's order encouraged vaccination
\flaw{without specifying voluntary or mandatory status}.
Another source described
\flaw{a COVID-19 vaccine mandate~for~service~members}.
&
Trump's September 20, 2020 order encouraged vaccination.
A Department of Defense document described a military mandate
and noted that
\good{public health responsibilities primarily lie with
state~and~local~governments}.
&
\\

\addlinespace[2pt]

\caseJudgmentIcon\hspace{0.4em}\textbf{Judgment}
&
Mandates applied to \flaw{specific groups},
while Trump focused on
\flaw{vaccine development and distribution}.
These are conflicting:
~\flaw{`\textsc{Conflicting/Cherry-picking}'}.~\xmark
&
Vaccination mandates primarily
\good{fall under state and local authority}.
No federal mandate or declaration of voluntary vaccination
for the general public was~identified:
~\good{\textit{\underline{`\textsc{Refuted}'}}}.~\cmark
&
\\

\addlinespace[2pt]
\midrule


\multicolumn{4}{@{}>{\columncolor{myblue}}p{\dimexpr1.05\textwidth-2\tabcolsep+0.3cm\relax}@{}}{%
\textbf{Claim 2, \textsc{Refuted}:}
Amy Coney Barrett, associate justice of the Supreme Court of the
United States, said during her Supreme Court confirmation hearing
that she was not using notes because she has seven kids and
is used to stupid questions.
}\\

\addlinespace[2pt]

\caseQueryIcon\hspace{0.4em}\textbf{Query}
&
\begin{minipage}[t]{\linewidth}
\raggedright
\par\hangindent=1.6em\hangafter=1\noindent
\makebox[1.6em][l]{\textcolor{blueText}{1.}}
``Amy Coney Barrett confirmation hearing notes seven~kids~stupid~questions'';

\par\hangindent=1.6em\hangafter=1\noindent
\makebox[1.6em][l]{\textcolor{blueText}{2.}}
``Amy Coney Barrett~confirmation~hearing~transcript''.\par
\end{minipage}
&
\begin{minipage}[t]{\linewidth}
\raggedright
\par\hangindent=1.6em\hangafter=1\noindent
\makebox[1.6em][l]{\textcolor{blueText}{1.}}
``Amy Coney Barrett confirmation hearing notes seven
kids~stupid~questions~transcript'';

\par\hangindent=1.6em\hangafter=1\noindent
\makebox[1.6em][l]{\textcolor{blueText}{2.}}
``Amy Coney Barrett confirmation hearing~\good{October~2020~quotes}'';

\par\hangindent=1.6em\hangafter=1\noindent
\makebox[1.6em][l]{\textcolor{blueText}{3.}}
``Amy Coney Barrett Supreme Court~confirmation~hearing~video''.\par
\end{minipage}
&
\multirow{3}{*}{%
\includegraphics[width=0.25\textwidth]{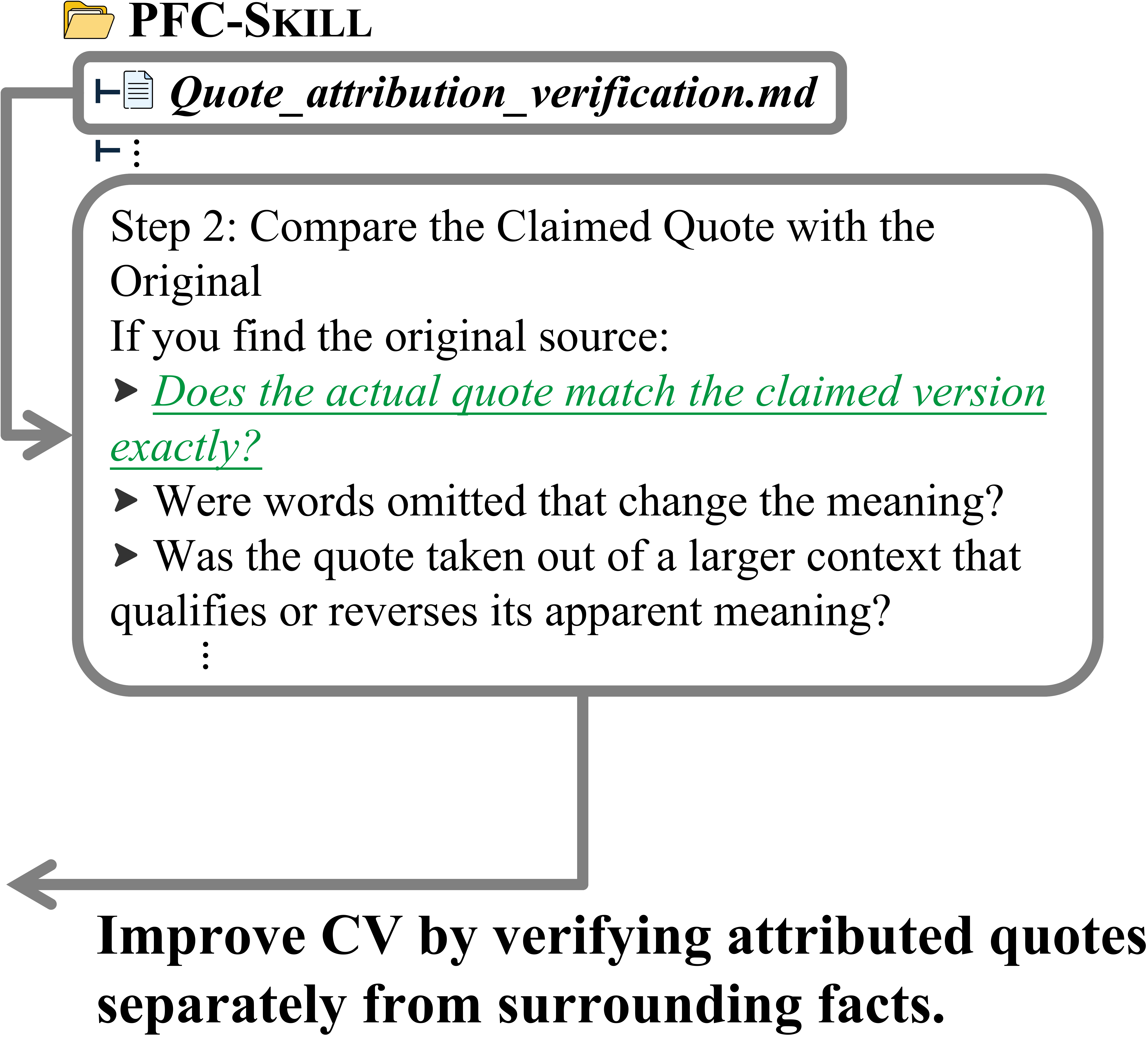}
}
\\

\addlinespace[2pt]

\caseEvidenceIcon\hspace{0.4em}\textbf{Evidence}
&
Barrett displayed a blank notepad,
showing she was~not~using~notes.
&
Barrett's blank notepad showed
\good{she used no notes};
it did not~verify~the~quote.
&
\\

\addlinespace[2pt]

\caseJudgmentIcon\hspace{0.4em}\textbf{Judgment}
&
The blank-notepad incident was taken to support the alleged remark,
\flaw{despite the lack of a verbatim~account}:
~\flaw{`\textsc{Supported}'}.~\xmark
&
Barrett was not using notes,
\good{\textit{\uline{but the attributed wording and context were~unverified}}}:
~\good{\textit{\underline{`\textsc{Refuted}'}}}.~\cmark
&
\\

\addlinespace[2pt]
\midrule


\multicolumn{4}{@{}>{\columncolor{myblue}}p{\dimexpr1.05\textwidth-2\tabcolsep+0.3cm\relax}@{}}{%
\textbf{Claim 3, \textsc{Supported}:}
52\% of Nigeria's current population lives in urban areas.
}\\

\addlinespace[2pt]

\caseQueryIcon\hspace{0.4em}\textbf{Query}
&
\begin{minipage}[t]{\linewidth}
\raggedright
\par\hangindent=1.6em\hangafter=1\noindent
\makebox[1.6em][l]{\textcolor{blueText}{1.}}
``Nigeria urban~population~percentage~2020'';

\par\hangindent=1.6em\hangafter=1\noindent
\makebox[1.6em][l]{\textcolor{blueText}{2.}}
``Nigeria urbanization rate~2020~World~Bank'';

\par\hangindent=1.6em\hangafter=1\noindent
\makebox[1.6em][l]{\textcolor{blueText}{3.}}
``Nigeria population urban areas~2020~census~data''.\par
\end{minipage}
&
\begin{minipage}[t]{\linewidth}
\raggedright
\par\hangindent=1.6em\hangafter=1\noindent
\makebox[1.6em][l]{\textcolor{blueText}{1.}}
``Nigeria urban population percentage~2020~World~Bank'';

\par\hangindent=1.6em\hangafter=1\noindent
\makebox[1.6em][l]{\textcolor{blueText}{2.}}
``Nigeria National Bureau of Statistics~urban~population~2020'';

\par\hangindent=1.6em\hangafter=1\noindent
\makebox[1.6em][l]{\textcolor{blueText}{3.}}
``United Nations World Urbanization~Prospects~Nigeria~2020'';

\par\hangindent=1.6em\hangafter=1\noindent
\makebox[1.6em][l]{\textcolor{blueText}{4.}}
``Nigeria urban population 2020
'National Bureau~of~Statistics'~report'';

\par\hangindent=1.6em\hangafter=1\noindent
\makebox[1.6em][l]{\textcolor{blueText}{5.}}
``'National Population Commission'
Nigeria~urban~percentage~2020'';

\par\hangindent=1.6em\hangafter=1\noindent
\makebox[1.6em][l]{\textcolor{blueText}{6.}}
``Nigeria demographic statistics 2020 urban
rural~split~official~report''.\par
\end{minipage}
&
\multirow{3}{*}{%
\includegraphics[width=0.25\textwidth]{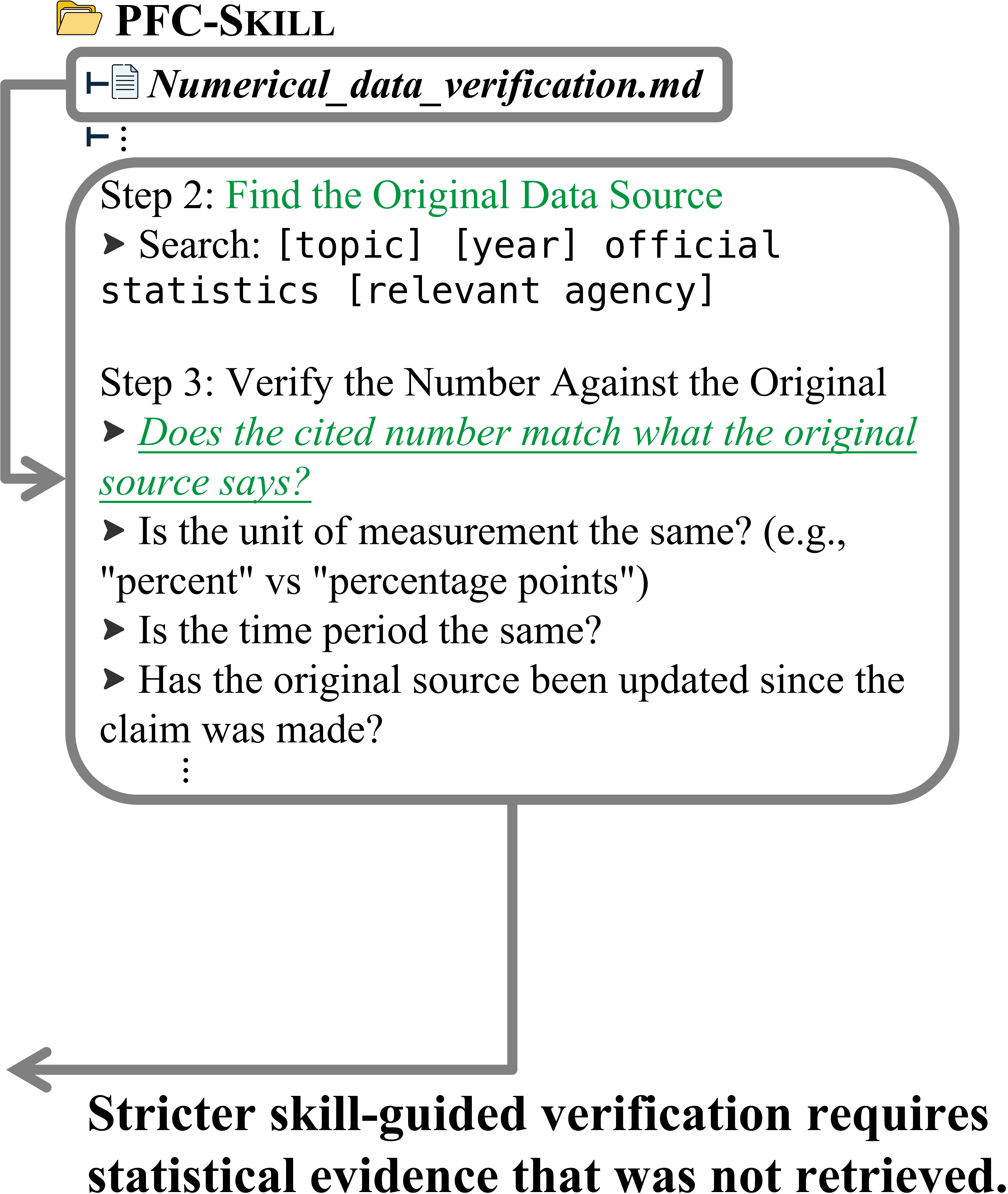}
}
\\

\addlinespace[2pt]

\caseEvidenceIcon\hspace{0.4em}\textbf{Evidence}
&
Sources gave a 53--54\% urban share,
including a 53.52\%~estimate~for~2022.
&
A later report gave 53.52\% for 2022.
Buhari's October 1, 2020 speech reported a 52\% urban share,
\flaw{without statistical data or sources~verifying~the~figure}.
&
\\

\addlinespace[2pt]

\caseJudgmentIcon\hspace{0.4em}\textbf{Judgment}
&
The 53.52\% estimate for 2022 was considered
\good{close enough to support the 52\% claim for 2020},
alongside World Bank and UN~data:
~\good{`\textsc{Supported}'}.~\cmark
&
Buhari stated the figure,
but later estimates did not directly verify it.
\flaw{\textit{\uline{In the absence of direct statistical verification of the 2020 figure}}},
the evidence was judged~insufficient:
~\flaw{`\textsc{NEI}'}.~\xmark
&
\\

\bottomrule

\end{tabular}
}
\end{table*}

\subsection{Case Studies and Error Analysis}\label{sec:case_study}
To better understand how \textsc{PFC-Skill} influences the fact-checking behaviors of LLM agents and leads to the observed improvements, we qualitatively analyze their query formulation during evidence retrieval (ER) and evidence assessment during claim verification (CV).
Table~\ref{tab:rq2_cases1} compares \textsc{SkillCheck} with the skill-free baseline DEFAME on three representative claims.
Claims~1 and~2 illustrate the benefits of skill guidance during ER and CV, respectively, while Claim~3 reveals an error introduced by the same guidance.

\textbf{In Claim~1, \textsc{PFC-Skill} improves ER by steering queries toward the scope and authority of vaccination policies.}
DEFAME retrieves a general encouragement of vaccination together with a military mandate and treats them as conflicting evidence.
\textsc{SkillCheck} instead searches for Trump's statements and the federal policy framework, surfacing the fact that public-health authority primarily rests with state and local governments.
This yields the correct veracity label, \textsc{Refuted}.
\textbf{In Claim~2, \textsc{PFC-Skill} improves CV by separating verification of an attributed quote from verification of surrounding facts.}
DEFAME takes the blank-notepad incident as sufficient support for the alleged remark.
\textsc{SkillCheck} confirms that Barrett used no notes but finds no primary source for the specific wording, changing the predicted veracity label from \textsc{Supported} to \textsc{Refuted}.
\emph{Together, these cases show that \textsc{PFC-Skill} strengthens both ER and CV by encouraging more precise query formulation and stricter scrutiny of retrieved evidence.}

To better understand the limitations of this guidance, we also examine Claim~3, where the same skill leads to an overly conservative error.
\textbf{In Claim~3, \textsc{SkillCheck} produces an error because stricter skill-guided verification demands precise statistical evidence that is not retrieved.}
DEFAME accepts an urban population estimate of 53--54\% as adequate support for the 52\% claim and correctly predicts \textsc{Supported}.
\textsc{SkillCheck} demands more direct statistical evidence for the precise 2020 figure; when no evidence supporting that year's figure is retrieved, it predicts \textsc{NEI}.
Although this decision is reasonable given the lack of precise evidence, the approach may be too conservative.

\section{Conclusion}
In this work, we introduce \textbf{\textsc{PFC-Skill}}, an agent skill distilled from AFP Fact Check's professional guidelines, and propose \textsc{SkillCheck}, a training-free AFC system that dynamically selects relevant skills to guide evidence retrieval and claim verification.
Experiments demonstrate that \textsc{SkillCheck} outperforms the previous state-of-the-art skill-free AFC system, DEFAME. Furthermore, \textsc{SkillCheck} with GPT-4o-Mini achieves higher AFC performance at lower inference cost than skill-free DEFAME with GPT-4o.
Variation in gains across LLM backbones also motivates future research in skill learning to enhance LLMs' ability to apply skills.
Ablation studies show that evidence retrieval guidance accounts for most gains, while claim verification guidance provides further improvements.
Case studies illustrate that skill guidance enables more targeted evidence retrieval and closer scrutiny during claim verification.

As the first work to demonstrate the value of distilling professional fact-checking expertise into reusable, actionable skills for LLM agents, our approach provides a practical pathway toward more capable and cost-efficient fact-checking systems without requiring additional model training or continuous human expert supervision. 
\textit{This supports ongoing efforts to foster a trustworthy information ecosystem for a responsible Web.}
In future work, we plan to incorporate additional skill sources from professional fact-checking organizations, reduce inference costs by investigating more efficient skill selection and context loading, and examine how these methods can generalize to multilingual scenarios.

\clearpage

\bibliography{main}
\bibliographystyle{ACM-Reference-Format}

\appendix

\begin{table*}[!t]
\centering
\caption{
Detailed descriptions of the reference files in \textsc{PFC-Skill} used in \textsc{SkillCheck}.
}
\label{tab:skill_details}
\resizebox{0.90\textwidth}{!}{

\begin{tabular}{p{0.41\textwidth} p{0.66\textwidth}}
\toprule

\textbf{Reference File}
&
\textbf{Reference Content}
\\

\midrule

\texttt{Claim\_decomposition.md}
&
Breaking complex claims into atomic, verifiable sub-questions; \newline
Claim type identification; \newline
Investigation prioritization.
\\
\cmidrule(lr){1-2}

\texttt{Search\_query\_strategies.md}
&
Concrete search query construction techniques by claim type; \newline
Iterative refinement strategies; \newline
Query anti-patterns.
\\
\cmidrule(lr){1-2}

\texttt{Source\_credibility\_evaluation.md}
&
4-tier source reliability hierarchy; \newline
Source evaluation checklist; \newline
Rules for handling denials, expert citations, and prior fact-checks.
\\
\cmidrule(lr){1-2}

\texttt{Cross\_verification\_methods.md}
&
Multi-source verification strategies, including source independence criteria, consistency checks, and handling conflicting evidence.
\\
\cmidrule(lr){1-2}

\texttt{Evidence\_sufficiency\_evaluation.md}
&
Criteria for evidence sufficiency; \newline
Evidence weight taxonomy; \newline
Handling absence of evidence; \newline
When to continue or stop retrieval.
\\
\cmidrule(lr){1-2}

\texttt{Contextual\_analysis.md}
&
Detecting out-of-context usage, cherry-picking, misleading framing, and omitted context; \newline
Distinguishing misleading claims from incomplete context.
\\
\cmidrule(lr){1-2}

\texttt{Numerical\_data\_verification.md}
&
Verification techniques for statistical and numerical claims, including tracing original sources, checking methodologies, and detecting manipulation patterns.
\\
\cmidrule(lr){1-2}

\texttt{Quote\_attribution\_verification.md}
&
Verifying the authenticity of attributed quotes; \newline
Detecting truncation, splicing, and fabrication; \newline
Searching for original transcripts.
\\
\cmidrule(lr){1-2}

\texttt{Confirmation\_bias\_countermeasures.md}
&
Bias mitigation strategies, including mandatory counter-evidence search, impartiality standards, and pre-verdict self-check procedures.
\\
\cmidrule(lr){1-2}

\texttt{Verdict\_reasoning\_logic.md}
&
Systematic reasoning from evidence to verdict, including evidence direction assessment, verdict strength calibration, and attribution-based justification.
\\

\bottomrule
\end{tabular}
}
\end{table*}

\FloatBarrier

\section{\textsc{PFC-Skill} Reference Files}\label{app:skill_details}

Table~\ref{tab:skill_details} provides detailed descriptions of the reference files in the unified \textsc{PFC-Skill}.

\section{Skill Incorporation Prompt}\label{app:prompt}

The agent reads \texttt{SKILL.md} and the reference catalog when selecting reference files. After selection, the contents of the selected files are loaded into its context for ER or CV.

\begin{tcolorbox}[
enhanced,
breakable,
sharp corners,
colback=gray!10,
colframe=black,
boxrule=0.5pt,
left=3pt,
right=3pt,
top=3pt,
bottom=3pt,
before={\par\addvspace{3pt}\noindent},
after={\par\addvspace{8pt}}
]
\small

\textbf{\# Instructions}

You are selecting reference guides for a fact-check.

\par\addvspace{0.3em}

\textbf{\#\# Current Stage}

\textbf{Stage}: [STAGE] 

[STAGE\_HINT]

\par\addvspace{0.3em}

Review the document below and the available reference files, then select the ones most relevant to this stage of the fact-check.

Output ONLY the file paths, one per line. If none are relevant, write ``None''.

\par\addvspace{0.3em}

\textbf{\#\# Document}

[DOC]

\par\addvspace{0.3em}

\textbf{\#\# Available References}

[SKILL\_CATALOG]

\par\addvspace{0.3em}

\textbf{\#\# Your Selected References}
\end{tcolorbox}

[STAGE]: Identifies the current stage: evidence retrieval (ER) or claim verification (CV).

[DOC]: Contains the claim and the accumulated fact-checking record.

[STAGE\_HINT]: Provides guidance on the intended behaviors at the current AFC stage.

[SKILL\_CATALOG]: Describes the available reference files in \textsc{PFC-Skill}.

\section{Definitions of the Veracity Labels}
\subsection{AVeriTeC}\label{app:label1}
AVeriTeC defines four veracity labels~\cite{averitec}. We use shortened names below. The label \textit{Not Enough Evidence} is denoted by \textsc{NEI}:
\begin{itemize}[leftmargin=*]
    \item \textbf{\textsc{Supported}}: the claim is supported by the available evidence.

    \item \textbf{\textsc{Refuted}}: the claim is contradicted by the available evidence.

    \item \textbf{\textsc{NEI}} (\textit{Not Enough Evidence}): the available evidence is insufficient to verify or refute the claim.

    \item \textbf{\textsc{Conflicting/Cherry-picking}}: the claim is misleading due to conflicting evidence or omission of important context.
\end{itemize}

\subsection{\texorpdfstring{ClaimReview2024+$_{\text{date}}$}{ClaimReview2024+ (date)}}\label{app:label2}
ClaimReview2024+$_{\text{date}}$ uses four veracity labels defined in \cite{defame}:
\begin{itemize}[leftmargin=*]
    \item \textbf{\textsc{Supported}}: the claim is accurate based on the available evidence.

    \item \textbf{\textsc{Refuted}}: the available evidence contradicts the claim.

    \item \textbf{\textsc{Misleading}}: the claim is misleading or requires additional context.

    \item \textbf{\textsc{NEI}} (\textit{Not Enough Information}): there is insufficient evidence to verify the claim.
\end{itemize}

\end{document}